\documentclass[runningheads]{llncs}
\usepackage[T1]{fontenc}
\usepackage{graphicx}
\usepackage{multicol}
\usepackage{algorithm}
\usepackage{soul}
\usepackage{url}
\usepackage{booktabs}
\usepackage{array}
\usepackage{color, colortbl}
\usepackage{multirow}
\usepackage{autobreak}
\usepackage{amssymb}
\usepackage{makecell}
\usepackage{pifont}
\usepackage{fontawesome}
\usepackage{bm}
\usepackage{wrapfig}
\usepackage{relsize}
\usepackage{tabularx}
\usepackage{mathrsfs}
\usepackage{lipsum}
\usepackage{listings}
\usepackage{setspace}
\usepackage{algpseudocode}

\usepackage{hyperref}

\newcommand{\ie}{\textit{i.e.,} }

\begin{document}

%
\title{Improving RL Exploration for LLM Reasoning\\through Retrospective Replay}

\author{Shihan Dou\inst{1} \and
Muling Wu\inst{1} \and
Jingwen Xu\inst{1} \and
Rui Zheng\inst{1} \and \\
Tao Gui\inst{1}\thanks{Corresponding author.} \and
Qi Zhang\inst{1} \and
Xuanjing Huang\inst{1}
}
\authorrunning{Dou et al.}
%
\institute{
Fudan University, Shanghai, China \\
\email{shdou21@m.fudan.edu.cn, tgui@fudan.edu.cn}
}


%
\maketitle              
\begin{abstract}

Reinforcement learning (RL) has increasingly become a pivotal technique in the post-training of large language models (LLMs).
The effective exploration of the output space is essential for the success of RL. 
We observe that for complex problems, during the early stages of training, the model exhibits strong exploratory capabilities and can identify promising solution ideas. 
However, its limited capability at this stage prevents it from successfully solving these problems.
The early suppression of these potentially valuable solution ideas by the policy gradient hinders the model's ability to revisit and re-explore these ideas later. 
Consequently, although the LLM's capabilities improve in the later stages of training, it still struggles to effectively address these complex problems.
To address this exploration issue, we propose a novel algorithm named \textbf{R}etrospective \textbf{R}eplay-based Reinforcement \textbf{L}earning (\textbf{RRL}), which introduces a dynamic replay mechanism throughout the training process. 
RRL enables the model to revisit promising states identified in the early stages, thereby improving its efficiency and effectiveness in exploration. 
To evaluate the effectiveness of RRL, we conduct extensive experiments on complex reasoning tasks, including mathematical reasoning and code generation, and general dialogue tasks. 
The results indicate that RRL maintains high exploration efficiency throughout the training period, significantly enhancing the effectiveness of RL in optimizing LLMs for complicated reasoning tasks.
Moreover, it also improves the performance of RLHF, making the model both safer and more helpful.

\keywords{LLMs \and Reinforcement Learning \and Reasoning}
\end{abstract}

\section{Introduction}

Recently, reinforcement learning (RL) has progressively emerged as a crucial approach for the post-training of large language models (LLMs), enhancing their capabilities in complex reasoning tasks such as mathematical reasoning \cite{bi2024deepseek,Hendrycks2021MeasuringMP} and code generation \cite{zheng2023delve,wang2024secrets}. 
The core concept of RL involves balancing \textbf{exploration} and \textbf{exploitation} \cite{Overcoming2018,exploration1992}.
It allows the model to extensively explore the output space and then optimize through the policy gradient by learning from the solution paths discovered during this exploration \cite{li2017deep}.
Consequently, efficient exploration and the discovery of correct trajectories are crucial for the success of RL in LLMs \cite{yang2021exploration,Ecoffet2020FirstRT}.

However, we observe that for complex problems, although the model exhibits strong exploratory capabilities and can identify promising solution ideas (\ie states) during the early stages of training, its weaker capabilities at this stage prevent it from generating correct solution paths. 
This results in the policy gradient \cite{sutton1999policy} gradually optimizing the model to abandon these potentially valuable solution ideas. 
As training progresses, the model's capabilities improve, but when faced with these complex problems again, the adverse effects of early policy gradient optimizations become apparent. 
The suppression of these potentially correct problem-solving ideas hinders the model's ability to re-explore them, as shown in Figure~\ref{fig:motivation}. 
In other words, while the policy gradient enhances the model's capabilities through ``exploitation'', it also gradually reduces its ``exploration'' within the output space. 
This leads the model to forget these previously explored promising solution ideas, still struggling to find correct solution paths for complex problems.

To address this issue, we propose a novel RL algorithm, \textbf{R}etrospective \textbf{R}eplay-based Reinforcement \textbf{L}earning (\textbf{RRL}), which introduces a dynamic replay mechanism throughout the training process to enhance the exploration efficiency and effectiveness of reinforcement learning. 
RRL utilizes the value model to identify promising states discovered in the early stages of training and these states are stored in a buffer. 
As training progresses and the model's capability increases while its exploration decreases, we replay these potentially useful states, enabling the model to continue exploring them. 
At this stage, the more capable model is expected to retry these promising solution ideas, eventually solving complex problems to improve the effectiveness of RL training. 
Moreover, RRL also stores and replays intermediate states of canonical solutions for these problems to further enhance exploration of difficult problems.

To validate the effectiveness of RRL, we conduct extensive experiments across a wide range of tasks, including mathematical reasoning \cite{Cobbe2021TrainingVT,Hendrycks2021MeasuringMP}, code generation \cite{dou2024stepcoder,apps}, and general dialogue\cite{bai2022training}. 
The results show that RRL significantly enhances the model's exploration efficiency and effectiveness in challenging problems, improving ability on these tasks.
For instance, through RRL, LLMs enhance the ability of code generation and avoid syntax and semantic errors.
Moreover, we show that it also improves the Reinforcement Learning from Human Feedback (RLHF), enhancing human alignment and making the model both safer and more helpful. 
Overall, our main contributions are summarized as follows:
\begin{enumerate}
    \item We reveal that within the current RL algorithm framework, the model’s exploration of the output space gradually decreases throughout the training process. 
    This reduction in exploration prevents the model from further discovering correct solution paths for complex problems. 
    \item We introduce Retrospective Replay-based Reinforcement Learning (RRL), which stores promising states identified in the early stages and replays them at appropriate steps, to maintain high exploration capabilities in the whole RL training phase. 
    \item Experimental results show that RRL preserves the model's exploration abilities throughout the training process, which increases the ability to solve complex problems. Moreover, it also effectively enhances the performance of RLHF.
\end{enumerate}

\section{Retrospective Replay-based Reinforcement Learning}

In this section, we first clearly illustrate the exploration challenges of reinforcement learning in LLM reasoning, and then elaborate on the methodological details of RRL.

\subsection{Motivation: exploration challenges in reinforcement learning}
\label{sec:motivation}

\begin{wrapfigure}{r}{0.4\textwidth} 
  \centering
  \vspace{-3em}
  \includegraphics[width=0.4\textwidth]{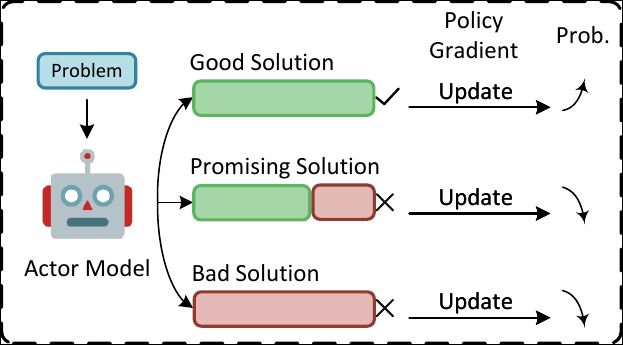} 
  \vspace{-2em}
  \caption{Policy gradient suppresses the generated solutions that have the correct idea but contain slight flaws.}
\label{fig:motivation}
\vspace{-1.5em}
\end{wrapfigure}

We first formally define the sampling and optimization process of the reinforcement learning algorithm in the context of LLMs.
Suppose the model $\pi_\theta$ is trained on a reasoning task such as code generation, where $q$ denotes the programming problem. 
It first needs to generate solutions based on this problem, that is, rollout trajectories. 
We assume the generated solution $y$ with $n$ tokens can be represented as $(y_1, y_2, \ldots, y_n)$. 
The state $s_i$ and action $a_i$ at the $i$-th timestep can be defined as $[q; y_{1:{i-1}}]$ and $y_i$, respectively.
Assume $r$ denotes the reward assigned according to the correctness of the solution. 
The model is then fine-tuned using the policy gradient \cite{sutton1999policy} on this trajectory:
\begin{equation}
\label{eq:pg}
\small
    \max_\theta \ E_{(q,y) \sim D_{\pi_\theta}} \left[ \sum_i A^{i} \log P(a_i \mid s_i; \theta) \right],
\end{equation}
where $P(a_i \mid s_i)$ is the probability of the policy model generating token $y_i$ according to the sequence $[q; y_{1:{i-1}}]$, and $A^i$ is the advantage computed by the generalized advantage estimator (GAE) \cite{schulman2015high} from $r$ to reduce the variability of predictions.
The policy model needs to rollout the correct solution to obtain the positive reward to optimize. 
Assume the token number of this correct solution is $n^*$, the probability of the policy generating this solution can be denoted as $P$, which can be written as:
\begin{equation}
\small
    P = \prod_{i=1}^{n^*} p(a_i \mid s_i).
\end{equation}
Complex programming problems often involve more conditional statements, making the solution more complex and longer, which causes $P$ to potentially decrease exponentially.
This means that, although the initial policy model may explore promising states, individual erroneous actions could lead to the overall solution being wrong and the reward being negative. At this time, the policy gradient gradually suppresses these promising but ultimately incorrect solution paths through Equation~\ref{eq:pg}. 
Consequently, the policy model is gradually discouraged from exploring these states, leading to the forgetting of potentially valuable states, as illustrated in Figure~\ref{fig:motivation}.

\begin{figure}[htbp]
    \centering
    \includegraphics[width=0.95\textwidth]{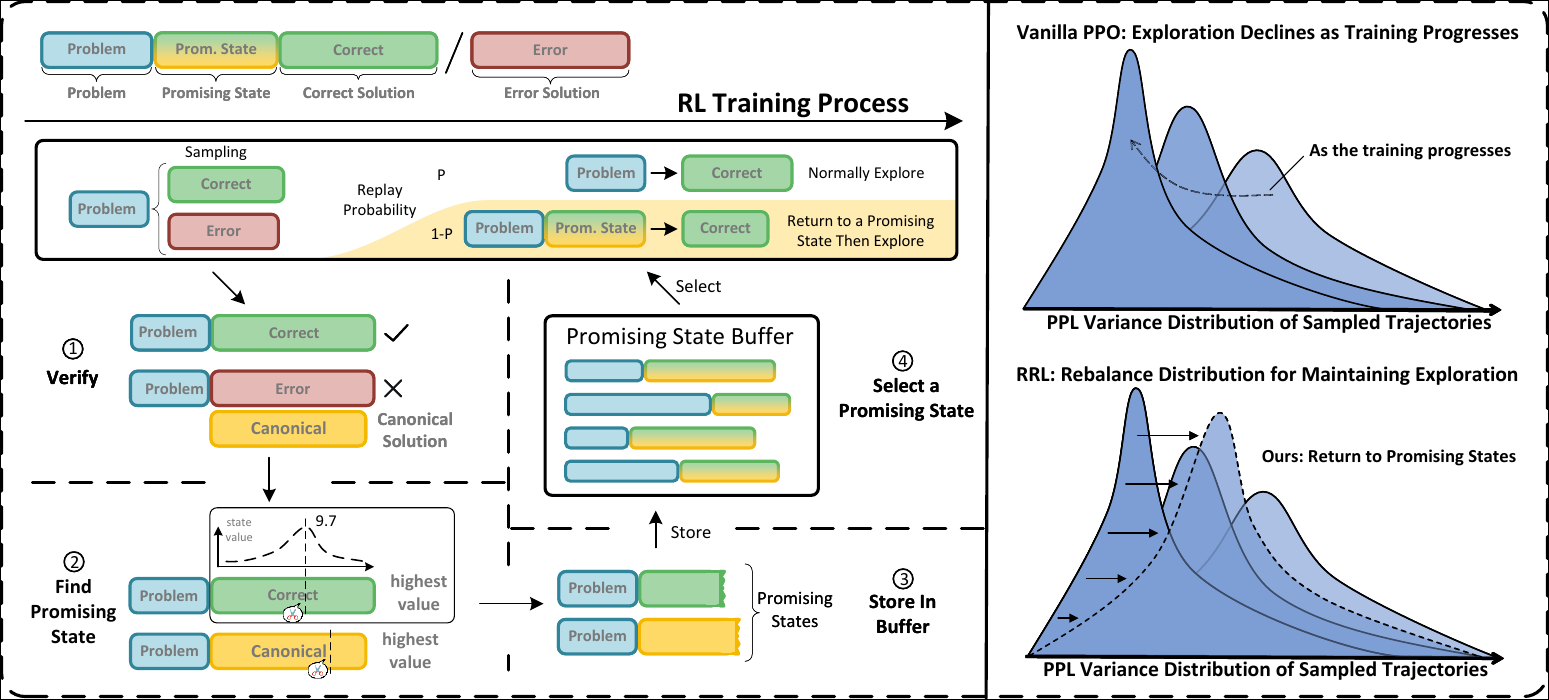}
    \caption{
    \textbf{Left:} Training process of RRL.
    \textbf{Right:} Distribution of the PPL variance of the responses obtained by performing multiple samplings for the same query.}
    \label{fig:pipeline}
    \vspace{-1em}
\end{figure}

Moreover, we have also observed that the policy gradient can reduce the diversity of actions taken by the policy model as RL training progresses, as shown in Figure~\ref{fig:pipeline}, which is consistent with findings in other work \cite{kirk2023understanding}. 
This further reduces the model's exploration of the output space. 
Due to these two reasons, the stronger model in the later stages of training cannot revisit the promising states explored by the weaker model in the early stages, leading to missed opportunities to solve these complex problems.
In summary, while the policy gradient improves the model's exploitative capabilities, it inadvertently reduces its exploratory capabilities over time. 
This dual effect hinders the model's ability to discover and utilize promising states, especially when faced with complex problems.

\subsection{Improving RL Exploration through retrospective replay}

We introduce a replay mechanism to enable the policy model to revisit promising states and continue exploring them to enhance their exploration ability.
In reinforcement learning, the value function aims to estimate the expected total reward starting from a given state and following the policy model. 
In other words, it evaluates the quality of the text segments (\ie states) generated by the model.
We utilize the value model to identify promising states explored from the early stages of training. 
Specifically, for a given problem, when the model generates a solution, we input this solution into the value model to find the state with the highest value and then add it to a buffer for replay: $\text{buffer} \leftarrow \arg\max_{s_i} V(s_i)$, where $V(s_i)$ represents the value of state $s_i$ as estimated by the value model.

However, since generated solutions may contain errors, the promising states identified might also include errors. 
To address this, we also input the canonical solution for the problem into the value model to find the highest value state within it and add this state to the buffer as well.
We control the exploration strategy with a replay coefficient, determining whether the policy model should start exploring from the original problem or these promising states.
As training progresses, the probability of starting exploration from promising states increases.
This process ensures that the model can also explore valuable states as its capabilities improve.
In practice, we utilize the proximal policy Optimization (PPO) algorithm \cite{schulman2017proximal} to optimize the policy model.
When starting exploration from intermediate states, we calculate the loss of the policy model while masking the token positions of states to prevent over-optimization.
The optimization objectives of the policy model for both cases can be uniformly by maximizing the function as follows: 
\begin{equation}
\label{eq:vanilla-rl}
\small
    \text{Objective}(\theta) = E_{(s, \tau) \sim D_{\pi_\theta}} [r - \beta \log (\pi_\theta (\tau | s)) / \pi^{\text{ref}} (\tau | s) ],
\end{equation}
where the $s$ and $\tau$ denote the starting state and the policy model generated parts.
We also observe that the value model may provide unreliable estimates in the early stages of training. 
So RRL only begins to find promising states after the loss of the value model has stabilized.
We also design an exit mechanism. 
When states are successfully resolved, they are removed from the buffer. 
Each state in the buffer also has a counter, and the model is more likely to choose states that have not been explored.
The maximum capacity of the buffer for each problem is set to five. 
When the buffer is full, the state with the highest counter is removed.
The full algorithm is detailed in Algorithm~\ref{alg}.

\begin{algorithm*}
\caption{Retrospective Replay-based Reinforcement Learning (RRL)}
\label{alg:algorithm1}
\begin{algorithmic}[1]
\Require policy model $\pi_\theta$, value model $V$, replay coefficient $\beta$
\Require training dataset $D=\{(x_i,\hat{y_i}) \mid 1 \le i \le n\}$, buffer $B$, replay dataset $D_r$
\For{each step}
    \State Calculate the probability $p$ of sampling from $B$: $p = \beta * (\text{Step} / \text{Step num of one epoch})$ if $\text{Epoch} = 1$ else $\beta$
    \State Sample a problem $q$ from the training dataset
    \If{normal exploration} \Comment{In this case, $s$ is the problem $q$}
        \State Generate solution $\tau$ for $q$ using $\pi_\theta$
        \State Find promising state $s_1=\arg\max_{s_i} V(s_i)$ in $y$, and add $s_1$ to $B$
        \State If $\tau$ is wrong, also find promising state $s_2$ from $\hat{y}$, and add $s_2$ to $B$
    \Else \Comment{Now the policy needs to generate a completion from a state}
        \State Select a state from $B$ according to $q$ and generate completion $\tau$
        \State Calculate reward $r$ based on correctness
        \State Increment the counter for the state
        \If{the solution is correct}
           Remove the state from the buffer
        \EndIf
    \EndIf
    \State Optimize the policy model and value model using PPO
\EndFor
\end{algorithmic}
\label{alg}
\end{algorithm*}

\section{Experiment}
To evaluate the effectiveness of RRL, we conduct extensive experiments on three tasks, including code generation, math reasoning, and general dialogue.

\subsection{Setup}

\textbf{Datasets, baselines, and evaluation.}
For the code generation task, we chose the highly challenging dataset, APPS+\cite{dou2024stepcoder}. 
We select the Deeseek-Coder-Instruct-6.7B \cite{Guo2024DeepSeekCoderWT} as our foundation model.
For the math task, we selected the two most widely used datasets, GSM8K\cite{Cobbe2021TrainingVT} and MATH\cite{Hendrycks2021MeasuringMP}. 
We use Deepseek-Math-7B \cite{Shao2024DeepSeekMathPT} as the foundation model.
To further evaluate RRL's effectiveness in RLHF, we also conduct experiments on the Anthropic's HH-RLHF \cite{DBLP:journals/corr/abs-2204-05862} dataset.
We utilize Llama-2-7B \cite{touvron2023llama} as the foundation model.
We also supervised fine-tuning these three LLMs on datasets, following previous work \cite{dou2024stepcoder,wang2024secrets,xi2024training}.
We compare our proposed method to a wide range of open-source LLMs including these three foundation models, WizardCoder-Python-V1.0 \cite{luo2023wizardcoder}, CodeLlama-Instruct \cite{roziere2023codellama}, and StarCoder \cite{li2023starcoder}.
We also compare to other baselines including PPO, PPOCoder \cite{shojaee2023execution}, RLTF \cite{liu2023rltf}, and GSI \cite{Ding2024MitigatingTN}.
For these three tasks, we evaluate all models following previous work \cite{dou2024stepcoder,wang2024secrets,xi2024training}.

\textbf{Implementation.}
During the SFT phase, we adopt a learning rate set at $2e^{-5}$, conduct training for one epoch, and employ a warm-up period of $0.1$ epochs. 
In the PPO and RRL training phase, we employ a learning rate of $5e^{-7}$ for the policy model and $1.5e^{-6}$ for the critic model.
The sampling temperature is set to $0.8$, top-p is set to $0.9$, the maximum output token length is set to $2048$ and the token-level KL penalty coefficient $\beta$ is set to $0.001$.

\begin{table*}[htbp]
\small
  \centering
  \begin{spacing}{0.8}
    \setlength{\tabcolsep}{0.4mm}{
\caption{Pass@1 results on the APPS+ dataset. CS and PGS denote canonical solutions and policy-generated solutions, respectively. RRL-based enhances LLM's exploration and helps them solve these complex problems.
}
\begin{tabular}{lc|ccc|c}
\toprule
\toprule
\multicolumn{1}{c}{\multirow{2}[2]{*}{\textbf{Models}}} & \multicolumn{1}{c|}{\multirow{2}[2]{*}{\textbf{Size}}} & \multicolumn{4}{c}{\textbf{APPS+}} \\
      &       & \multicolumn{1}{c}{\textbf{Introductory}} & \multicolumn{1}{c}{\textbf{Interview}} & \multicolumn{1}{c|}{\textbf{Competition}} & \multicolumn{1}{c}{\textbf{Overall}} \\
\midrule
StarCoder & 15.6B    & 6.3\%     & 4.1\%     & 0.7\%     &  4.7\% \\
CodeLlama-Instruct & 13B    & 33.3 \%    & 11.0\%     & 1.4\%     & 18.7\% \\
WizardCoder-Python  & 13B    & 39.7\%     & 15.1\%     & 4.3\%     & 23.6\% \\
DeepSeek-Coder-Instruct  & 6.7B    & 49.4\%     & 18.7\%     & 3.6\%   & 29.2\% \\
SFT on APPS+ & 6.7B    & \textbf{50.1\%}     & \textbf{19.0\%}   & \textbf{6.4\%}   & \textbf{29.8\%} \\
\midrule
\multicolumn{6}{c}{RL-based LLMs (Using DeepSeek-Coder-Instruct-6.7B as the backbone)} \\
\midrule
Vanilla PPO & 6.7B    & 53.7\%     & 20.1\%     & 5.0\%     & 31.7\% \\
PPOCoder & 6.7B    & 54.4\%     & 20.3\%     & 6.4\%     & 32.1\% \\
RLTF  & 6.7B    & 55.1\%     & 20.8\%     & 6.4\%     & 32.7\% \\
\midrule
\textbf{RRL} & 6.7B    & \textbf{57.3\%}     & \textbf{23.5\%}     & \textbf{7.9\%}     & \textbf{35.2\%} \\
\quad \textbf{States only from CS} & 6.7B    & 57.1\%     & 		22.6\% & 6.4\%      & 34.5\% \\
\quad \textbf{States only from PGS} & 6.7B    &  56.9\%   &  22.3\%   &  5.7\%   & 34.2\% \\
\bottomrule
\bottomrule
\end{tabular} }%
\end{spacing}
\label{tab:main-results}%
\vspace{-1.5em}
\end{table*}%

\subsection{Main Results}

\textbf{Results on code generation.}
We first evaluate our method on the widely used dataset, APPS+ \cite{dou2024stepcoder}, as shown in Table \ref{tab:main-results}.
Results show that our method achieves significant improvement compared to direct imitation learning from the training dataset. Moreover, compared to other RL methods, including vanilla PPO, PPOCoder, and RLTF, RRL achieved the best performance, with improvements of 3.5\%, 3.1\%, and 2.5\%, respectively, suggesting its promising application prospects in the field of code generation. 
Compared to these baselines, our proposed method offers better exploration for more complicated programming problems.
RRL employs caching and replaying to enable the policy model to fully utilize its exploratory capabilities during the early stages of training to remedy the exploration in later training stages.

\textbf{Results on math reasoning.}
We then evaluate RRL on math reasoning tasks using the GSM8K and MATH datasets. 
Table \ref{tab:math-general} shows that RRL significantly outperforms other baselines, including the few-shot settings, imitation learning, and the state-of-the-art baseline GSI. 
Specifically, compared to Vanilla PPO and GSI, RRL achieves an improvement of 1.9 PPL and 0.6 PPL, respectively, on GSM8K. 
Moreover, it shows a significant performance improvement on the relatively more difficult MATH dataset compared to the baselines. 
Overall, our framework can address the exploration issues in RL, achieving better performance than traditional RL algorithms.


\begin{table*}[htbp]
\small
  \centering
  \begin{spacing}{0.8}
    \setlength{\tabcolsep}{0.8mm}{
\caption{Results on math reasoning \textbf{(left)} and general dialogue \textbf{(right)}.}
    \begin{tabular}{l|cc||c|ccc|c}
    \toprule
    \toprule
    \multicolumn{1}{c|}{\multirow{2}[4]{*}{\textbf{Task}}} & \multicolumn{2}{c||}{\textbf{Math Reasoning}} & \multicolumn{5}{c}{\textbf{Alignment (RLHF-HH)}} \\
\cmidrule{2-8}          & \textbf{GSM8K} & \textbf{Math} &   \textbf{Opponent}    & \textbf{Win} & \textbf{Tie} & \textbf{Lose} & \textbf{Win rate} \\
    \midrule
    Few-Shot & 60.1\% & 28.2\% &   -    &    -   &   -    &   -    & - \\
    SFT   & 60.7\% & 28.7\% & \textbf{RRL} vs SFT &   \textbf{66}    &    11   &   23    & \textbf{74.2\%} \\
    PPO   & 68.8\% & 33.3\% & \textbf{RRL} vs PPO &   \textbf{31}    &    44   &    25   & \textbf{55.4\%} \\
    GSI   &   70.1\%    &   30.4\%    &     -  &    -   &   -    &  -     & - \\
    \textbf{RRL}   &   \textbf{70.7\%}    &   \textbf{34.3\%}  & - &  -    &    -   &    -     & - \\
    \bottomrule
    \bottomrule
    \end{tabular} }%
\end{spacing}
\label{tab:math-general}
\vspace{-1.5em}
\end{table*}%

\textbf{Results on general dialogue.}
To further verify the applicability of RRL, we introduce RL into Reinforcement Learning from Human Feedback (RLHF) to optimize LLMs on general dialogue tasks.
The results in Table~\ref{tab:math-general} demonstrate the clear superiority of RRL compared to SFT and vanilla PPO. 
Using GPT-4 evaluation, the responses generated by RRL-optimized LLMs achieve a 74.2\% win rate over SFT models (\ie win rate = (our win) / (our win + our lose)). 
This highlights the effectiveness of RRL in the general dialogue task and can significantly enhance the performance of RLHF.

\subsection{Discussion}

\textbf{Ablation analysis.}
In RRL, we identify promising states from both the policy-generated solutions and canonical solutions. 
To further evaluate the effectiveness of these two approaches, we conduct ablation experiments, as shown in Table~\ref{tab:main-results}. 
The results reveal significant differences among these methods, with the combination of the two achieving the best performance. 
If we consistently use part of the canonical solution as a prefix for generation, the model tends to degenerate into imitation learning, leading to a decline in its exploration ability. 
On the other hand, another strategy involves using the generated solution as a cache for storage. 
If the generated solution is inherently incorrect, this error can occur in the states that the value model deems most valuable, thereby preventing the model from exploring correct solution paths. 
The results underscore that the combination of these two strategies can significantly enhance RL exploration, enabling the model to solve more difficult problems.

\begin{wrapfigure}{r}{0.4\textwidth} 
  \centering
  \vspace{-2em}
  \includegraphics[width=0.4\textwidth]{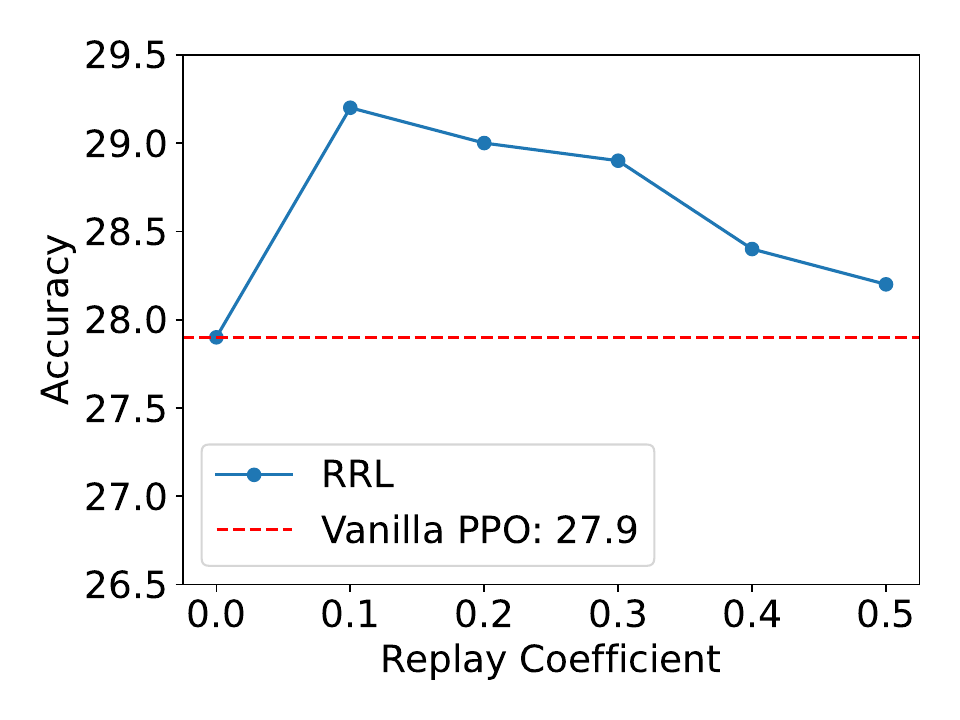} 
  \vspace{-2em}
  \caption{Sensitivity analysis.}
\label{fig:sen}
\vspace{-2em}
\end{wrapfigure}

\textbf{Sensitivity analysis.}
Compared to the vanilla PPO, we introduce an additional hyper-parameter, the replay coefficient $p$. 
During the sampling process, the policy has a probability of $p$ to generate a completion from a state in the buffer and a probability of $1 - p$ to directly generate a solution from the problem.
To effectively investigate the impact of the replay coefficient, we set it to six different values (i.e., 0, 0.1, 0.2, 0.3, 0.4, and 0.5) and conducted experiments.
Figure \ref{fig:sen} shows that, when the selection probability is set to 0.1, the model achieves optimal performance. 
Moreover, it shows that RRL under different replay coefficients can also bring improvements compared to the vanilla PPO, indicating our method can both achieve better performance under a range of settings.

\section{Related Work}

Recently, large language models have shown remarkable ability in reasoning and general conversation tasks \cite{christopoulou2022pangu,wang2024secrets,li2023starcoder,bi2024deepseek,Guo2024DeepSeekCoderWT,dou2024loramoe,dou2025evalearn,chai2024docfusion,dou2023towards,dou2024s,zang2025compression}.
Reinforcement learning has gradually become a core technology for enhancing the capabilities of large language models \cite{williams1992simple,shen2023loose,seed2025seed1,gao2024linear}.
Exploration is a critical concept in reinforcement learning \cite{Overcoming2018,exploration1992,5172767,ladosz2022exploration,8103164,dou2024stepcoder,wu2025progressive}. 
When faced with complex reasoning problems, large language models need to explore a vast generation space to identify correct solution paths and learn from these experiences \cite{dou2024stepcoder,zhao2023survey,kumar2025llm,plaat2024reasoning,huang2022towards}. 
Pre-trained language models often cannot be directly used for reinforcement learning because their exploration space is too large, which negatively impacts exploration efficiency \cite{guo2025deepseek,mercer2025brief}. 
Instruction fine-tuning can be considered a method to enhance exploration efficiency in reinforcement learning, as it narrows the exploration space of LLMs to a subspace, significantly reducing the complexity of exploration \cite{wang2024secrets,zheng2023delve,dou2024metarm}.
Moreover, some studies have demonstrated that few-shot-based in-context learning can also enable LLMs to follow instructions, thereby improving exploration efficiency in RL \cite{guo2025deepseek,wen2025light}.

Some work introduces process-supervised approaches to provide step-by-step supervised signals for complex reasoning tasks such as mathematical reasoning and code generation \cite{uesato2022solving}.
Process supervision can also be seen as an avenue to enhance exploration capabilities. 
Specifically, more complex problems often require generating longer solution trajectories \cite{dou2024s}. 
However, in tasks like complicated reasoning, the reward is sparse (\ie feedback based on the correctness of the final answer), making it difficult for models to explore correct paths and thus limiting their learning.
Process supervision methods alleviate this difficulty by training a reward model to supervise or using methods like Monte Carlo Tree Search (MCTS) to verify intermediate steps, thereby providing supervision for the process \cite{jiang2024rationalyst}. 
These methods partially mitigate the exploration challenges that models face in complex tasks.

A similar but entirely different concept is experience replay, which involves storing the experiences including states and actions in a buffer and replaying them to update policy and value functions \cite{schaul2015prioritized,zhang2017deeper}. 
In contrast, RRL replays previously explored potential intermediate states to continue exploration.
In addition, some approaches also introduce curriculum reinforcement learning, allowing models to start learning from simpler intermediate nodes and progressively tackle complete complex problems \cite{xi2024training,dou2024stepcoder,tao2024reverse}. 
Compared to these methods, our approach aims to maintain a high level of exploration throughout the training process. 
We are the first to reveal that LLMs, during exploration, may regrettably forget previously explored but not fully pursued potential solution ideas (\ie states) due to policy gradients.

\section{Limitations and Future Work}

Potential states may contain errors.
If the model continues to generate from these erroneous states, it will not find the correct trajectories. 
However, the model can still learn from these mistakes. 
Moreover, we also replay the promising intermediate states from the canonical solution of the problem to help the model continue generating from these correct intermediate states. 
In the future, we plan to find ways to identify and filter out erroneous promising states.
Additionally, we utilize the value model to identify promising intermediate states, as defined by the value function. 
The value model may be unstable and inaccurate in the early stages of training, leading to inaccurate results. 
Therefore, we discard the initial states. 
In the future, we plan to design more precise approaches to identify promising states.

\section{Conclusion}

In this paper, we reveal the negative impact that RL's policy gradient algorithm has on LLMs.
It causes LLMs to fail to re-explore potentially viable solution ideas (\ie states) for complex problems, limiting their exploration capacity in the later stages of RL training. 
We further propose a novel RL exploration framework named RRL. 
RRL introduces a novel replay mechanism to enhance the model's exploration throughout the training process. 
It utilizes the value model to identify potentially promising states explored in the early stages and dynamically replays these states when the LLM's exploration capability diminishes. 
RRL enables LLMs to continue reasoning and generating from these states, improving their ability to solve complex problems. 
Experimental results on tasks such as mathematical reasoning, code generation, and human alignment demonstrate that RRL maintains high exploration efficiency throughout the training period, significantly enhancing the effectiveness of RL in optimizing LLMs.

\section*{Acknowledgments}

The authors wish to thank the anonymous reviewers for their helpful comments. This work was partially funded by Guangdong S\&T Program 2024B0101050003.

%
%
%
\bibliographystyle{splncs04}
\bibliography{rrl}

\end{document}